\documentclass[11pt,letter]{article}
\usepackage[margin=1.2in]{geometry}
\usepackage[utf8]{inputenc}
\usepackage{multirow}
\usepackage{booktabs}
\usepackage{float}
\usepackage{amsmath}
\usepackage{amsfonts}
\usepackage{amssymb}
\usepackage{hyperref}
\usepackage{graphicx}
\usepackage{caption,sidecap}
\usepackage[labelformat=simple]{subcaption}
\usepackage{tikz}
\usetikzlibrary{shapes,arrows,positioning,calc}
\usepackage[round,authoryear]{natbib}

\DeclareMathOperator*{\argmin}{\arg\!\min}

\newtheorem{remark}{Remark}
\newtheorem{thm}{Theorem}

\newcommand\blfootnote[1]{%
  \begingroup
  \renewcommand\thefootnote{}\footnote{#1}%
  \addtocounter{footnote}{-1}%
  \endgroup
}

\title{Regularizing Recurrent Neural Networks via Sequence Mixup}
\author{Armin~Karamzade \and Amir~Najafi \and Seyed~Abolfazl~Motahari}

\begin{document}
\date{}
\maketitle
\begin{center}
Data Analytics Laboratory (DAL),\\
Computer Engineering Department,\\
Sharif University of Technology, Tehran, Iran
\vspace*{3mm}
\end{center}

\begin{abstract}
In this paper, we extend a class of celebrated regularization techniques originally proposed for feed-forward neural networks, namely Input Mixup \citep{zhang2017mixup} and Manifold Mixup \citep{verma2018manifold}, to the realm of Recurrent Neural Networks (RNN). Our proposed methods are easy to implement and have a low computational complexity, while leverage the performance of simple neural architectures in a variety of tasks. We have validated our claims through several experiments on real-world datasets, and also provide an asymptotic theoretical analysis to further investigate the properties and potential impacts of our proposed techniques. Applying sequence mixup to BiLSTM-CRF model \citep{huang2015bidirectional} to Named Entity Recognition task on CoNLL-2003 data \citep{sang2003introduction} has improved the F-1 score on the test stage and reduced the loss, considerably.
\blfootnote{Emails: \{karamzade,najafy\}@ce.sharif.edu,~motahari@sharif.edu}
\blfootnote{An implementation of our method is avaiable at \href{https://github.com/ArminKaramzade/SequenceMixup}{https://github.com/ArminKaramzade/SequenceMixup.}}
\end{abstract}


\section{Introduction}
\label{sec:intro}

Recurrent neural networks are the basis of the state-of-the-art models in natural language processing, including language modeling \citep{mikolov2011extensions}, machine translation \citep{cho2014learning} and named entity recognition \citep{lample2016neural}. It is needless to say that complex learning tasks require relatively large networks with millions of parameters to be accomplished. However, large neural networks need more data and/or strong regularization techniques to be trained successfully and avoid overfitting. Without the means to collect more data, which is the case in the majority of real-world problems, data augmentation and regularization methods are standard alternative practices to overcome this barrier.

Data augmentation in natural language processing is limited, and often task-specific \citep{kobayashi-2018-contextual, kafle-etal-2017-data}. On the other hand, adopting the same regularization methods that are originally proposed for feed-forward (non-recurrent) networks needs to be done with extra care to avoid hurting the network's information flow between consecutive time-steps. To overcome such limitations, we present Sequence Mixup: a set of training methods, regularization techniques, and data augmentation procedures for RNNs. Sequence Mixup can be considered as the RNN-generalization of input mixup \citep{zhang2017mixup} and manifold mixup \citep{verma2018manifold}, which are already introduced for feed-forward neural networks. Generally speaking, the core idea behind mixup strategies is to \textit{mix} training samples in the network's input or hidden layers, where by mix, we simply mean to consider random convex combinations of pairs of samples as alternatives for the actual training data points. Mixup in non-recurrent networks has led to smoother decision boundaries, more robustness to adversarial examples, and better generalization compared to many rival regularization methods \citep{zhang2017mixup, verma2018manifold}. Here, we extend input mixup to RNNs and also propose two variants of manifold mixup, namely Pre-Output Mixup (POM) and Through-Time Mixup (TTM), where mixing occurs in the hidden space of the RNN. POM and TTM differ from each other in the way information flow is passed from one time-step to the next.

\begin{figure*}[t]
	\centering
	\subfloat[]{\includegraphics[width=0.33\linewidth]{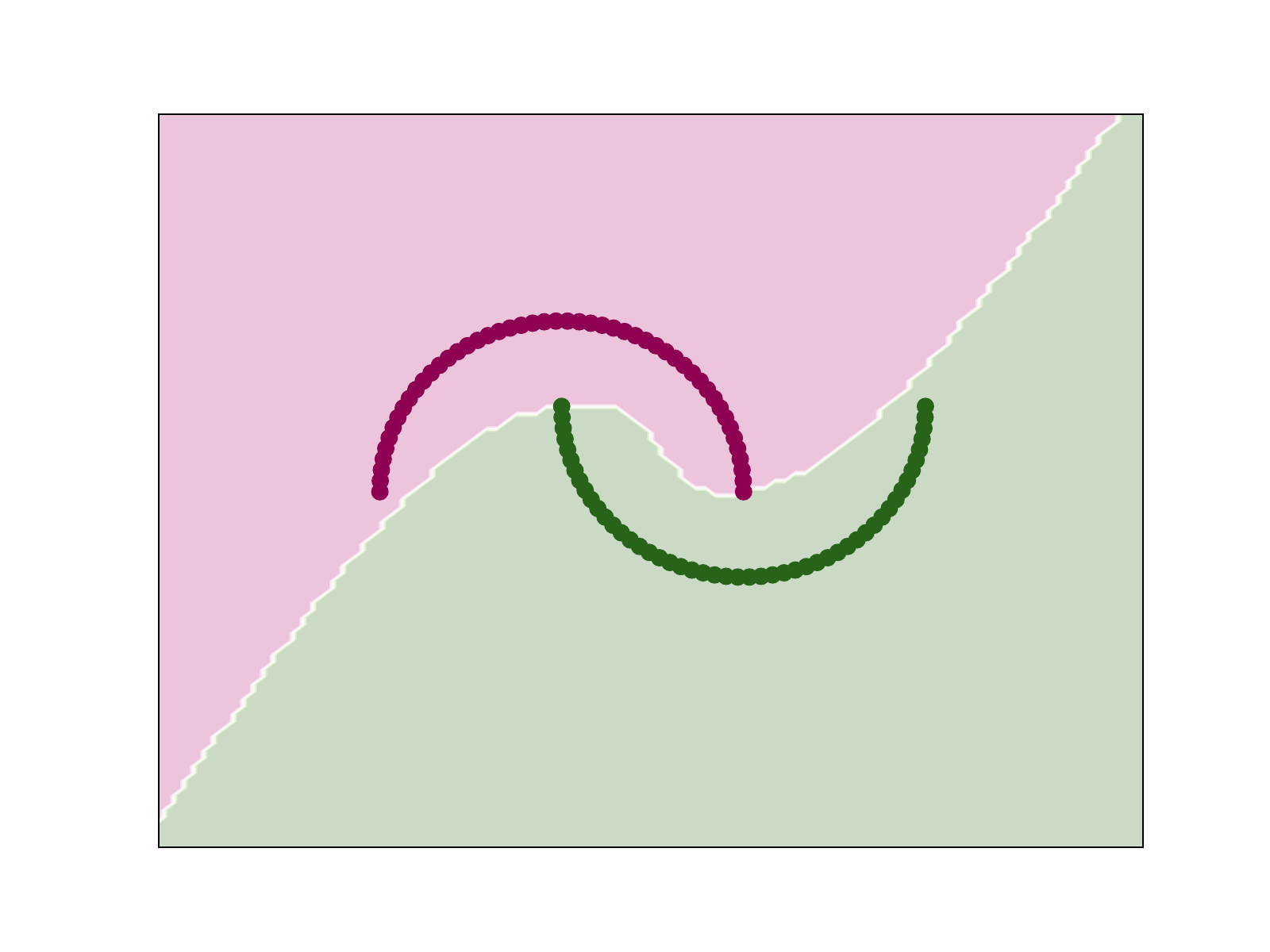}  \label{fig:true-db}}
	\subfloat[]{\includegraphics[width=0.33\linewidth]{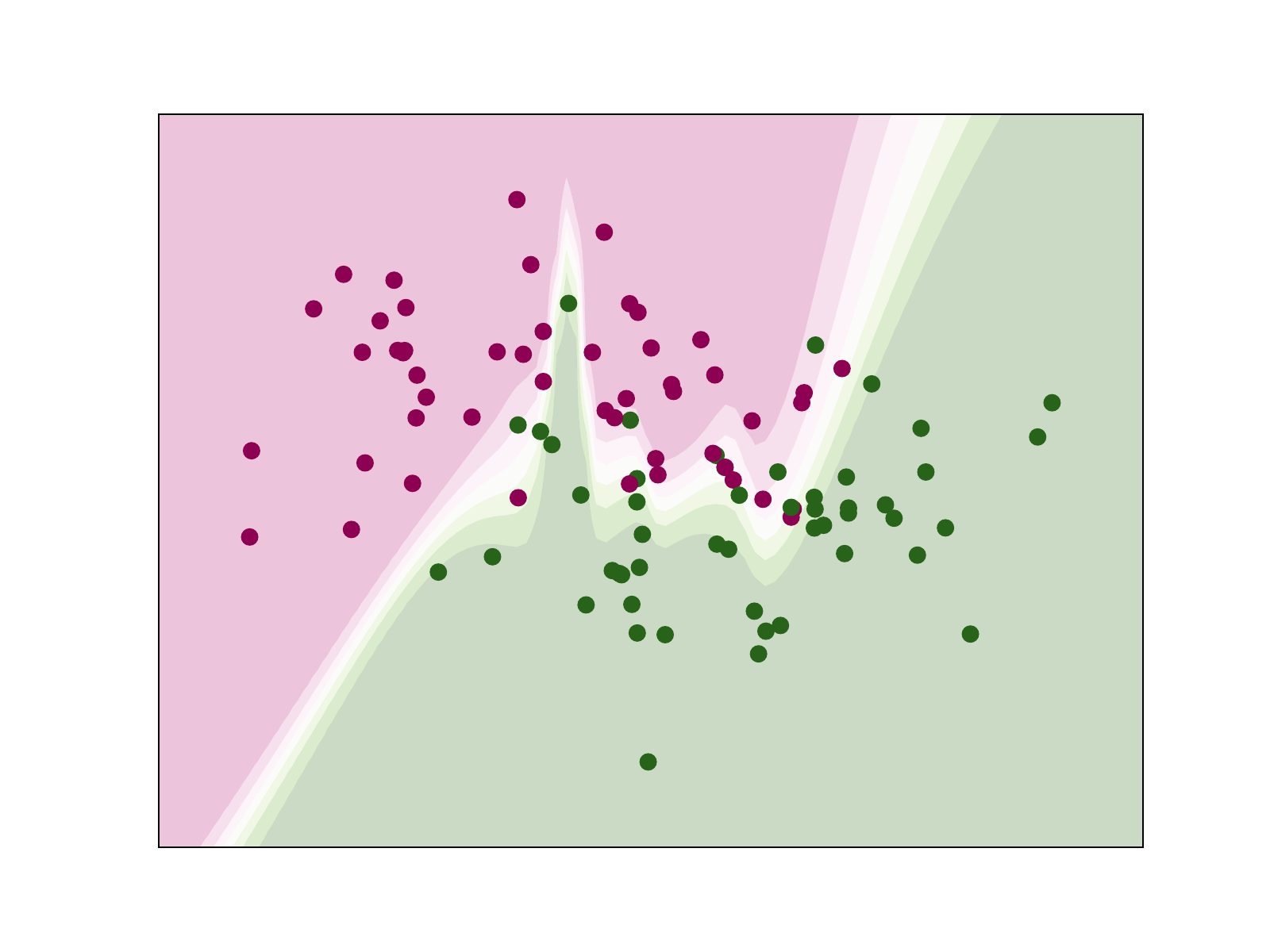}  \label{fig:normal-db}}
	\subfloat[]{\includegraphics[width=0.33\linewidth]{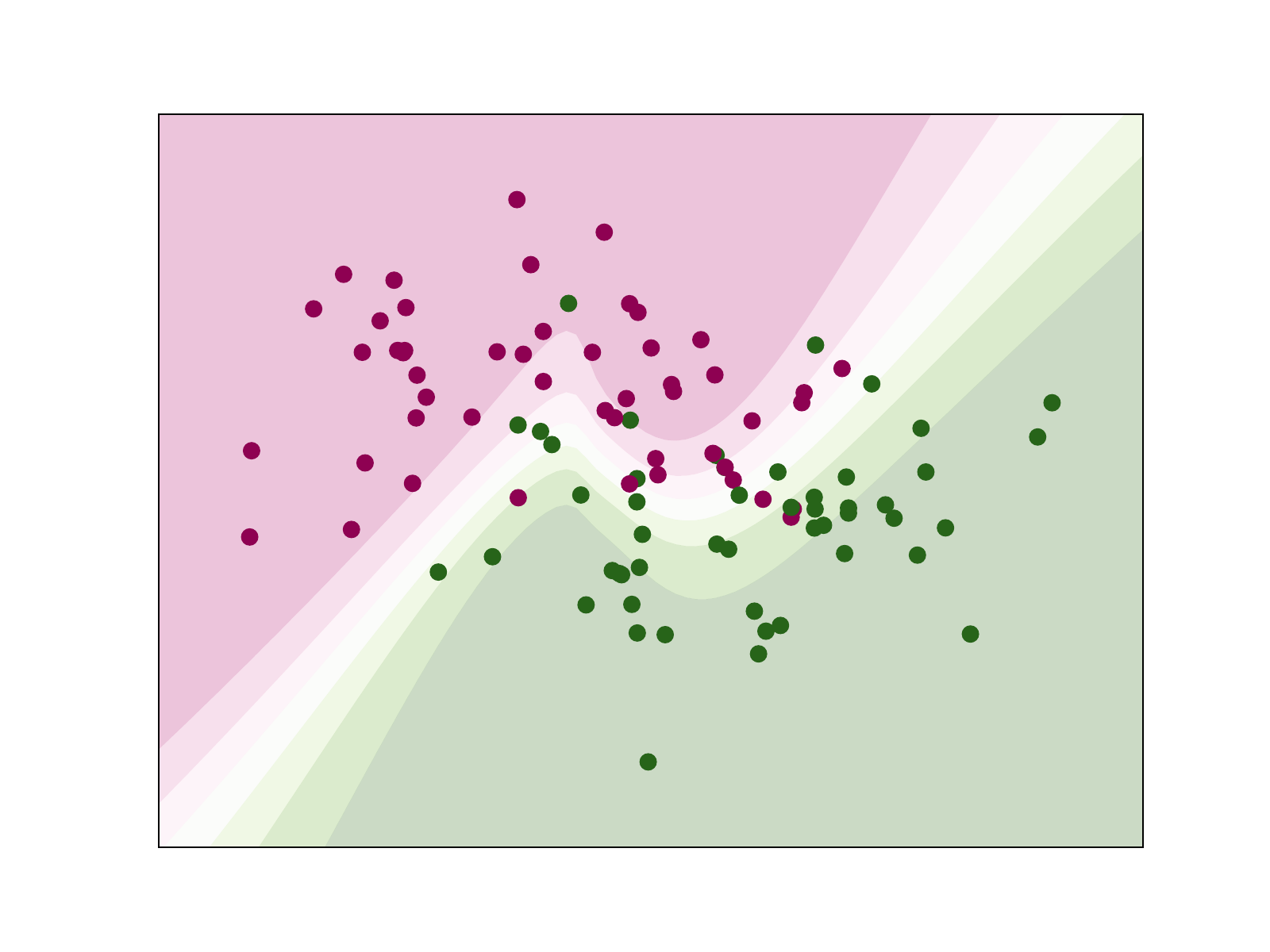}  \label{fig:mix-db}}
	\caption{\subref{fig:true-db} Original half-moons data and its learned decision boundary. Noise-corrupted half-moons and decision boundaries learned with \subref{fig:normal-db} regular training and \subref{fig:mix-db} Pre-Output Mixup (POM).}
	\label{fig:db}
\end{figure*}

In order to elucidate the effect of sequence mixup during the learning stage, consider the classification of half-moons data plotted in figure \ref{fig:true-db} with a simple two-timestep RNN. We have also added some levels of noise to the original data points to make the classification task more challenging. Figures \ref{fig:normal-db} and \ref{fig:mix-db} show the learned decision boundaries from noisy data via regular training and Pre-Output Mixup, respectively. As can be seen, mixup expands the margin between the classes and increases the decision boundary levels, which in turn renders a smoother decision boundary with less certainty about nearby cross-class samples. Intuitively speaking, this type of training creates artificial samples whose labels and hidden states are obtained from intermixing those of the original samples, in a respective manner. Based on our experiments, applying sequence mixup has improved both the test F-1 score and loss of BiLSTM-CRF model \citep{huang2015bidirectional} on CoNLL-03 data \citep{sang2003introduction} (Section \ref{subsec:evaluation}). 

We have also provided a theoretical analysis on the impact of our regularization techniques in the asymptotic regime where network widths become increasingly large, and learning rates become infinitesimally small. In a nutshell, our analysis reveals that as long as the number of hidden state neurons, which we denote by $H$ in this work, is less than the number of distinct classes in a classification problem, both POM and TTM cannot achieve a zero training error regardless of how large the training dataset is or how deep the neural networks become. Moreover, we show that as long as $H$ is less than twice the number of classes, the hidden-state generating section of the RNN acts as a memoryless unit and produces hidden states that are almost independent of previous time-steps. On the other hand, given that $H$ is chosen sufficiently large, both POM and TTM are able to divide the hidden representation space of the RNN into a set of orthogonal affine subspaces, where each subspace is an indicator of a unique class. We refer to this property as spectral compression of sequence mixup, which is a similar behaviour to that of manifold mixup for feed-forward networks.

The rest of the paper is organized as follows: Section \ref{sec:literature} reviews a number of related works to this problem. In Section \ref{sec:main}, we propose Sequence Mixup, describe its challenges and specifications in detail, and also present our theoretical analysis.  Section \ref{sec:experiments} is devoted to our experiments on real-world data. Finally, Section \ref{sec:conclusion} concludes the paper.

\section{Related Works}
\label{sec:literature}
Data augmentation is a popular technique for training large neural networks; It implicitly regularizes the model, which ultimately leads to a lower generalization error. There are several methods for data augmentation in a variety of areas such as computer vision and speech recognition. For example, cropping, translation, rotation, resizing, and flipping are prevailing techniques for creating new artificial images in computer vision tasks \citep{shorten}. Similarly, many successful methods such as SpecAugment \citep{park2019specaugment}, vocal tract length perturbation \citep{jaitly2013vocal} and the stochastic feature mapping approach of \cite{patent-20170040016} have been proposed for data augmentation in speech recognition. However, data augmentation for natural language processing tasks is more challenging. One cannot simply employ signal transformation methods used in image and speech processing tasks, since altering the order of words in a sentence can change both its synthetic and semantic meanings. Majority of existing data augmentation techniques for text either rely on replacing a word with its close alternatives, or are only applicable to specific domains. For instance, \cite{zhang2015character} used a thesaurus to obtain synonyms of a word for replacement, \cite{wang-yang-2015-thats} used $k$-nearest neighbors and cosine similarity to find similar words, and more recently \cite{kobayashi-2018-contextual} introduced contextual augmentation for extracting synonyms. Furthermore, \cite{sennrich-etal-2016-improving} employed a back translation technique for machine translation, and \cite{kafle-etal-2017-data} made use of a number of generative models for producing new questions in a visual question answering task.

Popular regularization techniques used in feed-forward networks usually fail to work on recurrent neural networks. For example, weight decay prevents RNNs from learning long-term dependencies \citep{pascanu2013difficulty}, and applying naive Dropout to RNNs causes deterioration of information flow through time-steps \citep{zaremba2014recurrent}. Various types of Dropout have been adopted for RNNs without hurting its memorization capability, e.g., \cite{zaremba2014recurrent} and \cite{pham2014dropout} applied Dropout only to non-recurrent sections of the network, \cite{gal2016dropout} used variational Dropout, \cite{krueger2016zoneout} introduced Zoneout, and  \cite{merity2017regularizing} applied the same dropout mask to recurrent weights. Apart from Dropout-based strategies, \cite{cooijmans2016recurrent} extended batch normalization \citep{ioffe2015batch} to RNNs for reducing internal covariate shift among time-steps. Also, \cite{dieng2018noisin} proposed Noisin for regularizing the network through injecting random noise into the hidden states of the RNN.

\section{Sequence Mixup}
\label{sec:main}

In this section, we build upon both Input Mixup \citep{zhang2017mixup} and Manifold Mixup \citep{verma2018manifold} regularizations from non-recurrent networks and make them applicable on RNNs. First, a set of necessary mathematical notations and definitions need to be established: For a finite set of time indices $\mathcal{T}=\left\{1,2,\ldots,T\right\}$, let us consider a recurrent neural network with input sequence $\boldsymbol{x}=\{\boldsymbol{x}_{t}\}$, corresponding one-hot label sequence $\boldsymbol{y}=\{\boldsymbol{y}_{t}\}$, set of hidden state vectors $\boldsymbol{h}_{t} = f(\boldsymbol{x}_{t}, \boldsymbol{h}_{t-1})$ and class predictions $\hat{\boldsymbol{y}}_{t}=g(\boldsymbol{h}_{t})$ for time-steps $t\in\mathcal{T}$. Here, functions $f$ and $g$ represent the state-generating and output-generating neural architectures in the RNN, respectively. Figure \ref{dig:simple} illustrates a block diagram which corresponds to the RNN described above. By a one-hot label $\boldsymbol{y}$, we simply mean a vector of dimension $C$, with $C$ denoting the number of distinct classes in a given classification task, where the dimension that corresponds to a particular class is set to $1$ while others are $0$. Also, let us assume the neural net that corresponds to function $f$ has $H$ output neurons, i.e., the hidden representation space is $H$-dimensional. Throughout the paper, and for any two vectors $\boldsymbol{z}$ and $\boldsymbol{z}'$ and coefficient $0\leq\lambda\leq1$, we denote $\lambda\boldsymbol{z} + (1-\lambda) \boldsymbol{z}^{\prime}$ by $\boldsymbol{z}^{\mathrm{mix}}(\lambda)$. Finally, let us denote $\boldsymbol{D}=\left\{\left(\boldsymbol{x}_i,\boldsymbol{y}_i\right)\vert~i=1,\ldots,n\right\}$ as the given training dataset including $n$ sample pairs.

\tikzset{
f/.style = {draw, fill=white, rectangle, minimum height=2em, minimum width=3em},
g/.style = {draw, fill=white, rectangle, minimum height=2em, minimum width=3em},
mix/.style= {draw, fill=white, circle, node distance=1.5cm},
tmp/.style  = {coordinate}
}
    
\begin{SCfigure}[40][t]
     \centering
     \begin{tikzpicture}[auto, node distance=1.5cm,>=latex',scale=0.85, every node/.style={scale=0.85}]
        \node [f, name=f1] (f1) {$f$};
        \node [f, right of=f1] (f2) [right=-2mm of f1] {$f$};
        \node [g, above=7mm of f1] (g1) {$g$};
        \node [g, above=7mm of f2] (g2) {$g$};
        \node [tmp, left of=f1, node distance=1.5cm] (tmp1) {};
        \node [tmp, right of=f2, node distance=1.9cm] (tmp11) {};
        \node [tmp, right of=f1, node distance=0.95cm] (tmp2) {};
        \node [tmp, right of=f2, node distance=0.95cm] (tmp22) {};
        \node [tmp, right of=tmp2, node distance=0.75cm] (tmp3) {};
        \node [tmp, below of=f1] (input1) [below=-5mm of f1] {};
        \node [tmp, above=5mm of g1] (out1) {};
        \draw [->] (g1) -- node[above=2mm of out1]{$\hat{\boldsymbol{y}}_t$} (out1);
        \node [tmp, above=5mm of g2] (out2) {};
        \draw [->] (g2) -- node[above=2mm of out2]{$\hat{\boldsymbol{y}}_{t+1}$} (out2);
        \draw [->] (input1) node[below]{$\boldsymbol{x}_t$} -- (f1);
        \node [tmp, below of=f2] (input2) [below=-5mm of f2] {};
        \draw [->] (input2) node[below]{$\boldsymbol{x}_{t+1}$} -- (f2);
        \draw[->] (f1) -- node[node distance=0.5cm, name=ht]{$\boldsymbol{h}_t$} (g1);
        \draw[->] (f2) -- node[node distance=0.5cm, name=ht1]{$\boldsymbol{h}_{t+1}$} (g2);
        \draw [->] (tmp1) node [left]{$\cdots~\boldsymbol{h}_{t-1}$} -- (f1);
        \draw [->] (ht) -| (tmp2) |-  node[right of=tmp3, node distance=1.05cm]{} (f2);
        \draw [->] (ht1) -| (tmp22) |-  node[right of=tmp3, node distance=1.8cm]{$\boldsymbol{h}_{t+1}~\cdots$} (tmp11);
    \end{tikzpicture}
    \caption{Block diagram of an RNN with hidden-generating function $f$ and output-generating function $g$. Functions $f$ and $g$ correspond to two distinct (and generally) deep neural networks. In a standard RNN, different time-steps have identical neural networks.}
    \label{dig:simple}
\end{SCfigure}
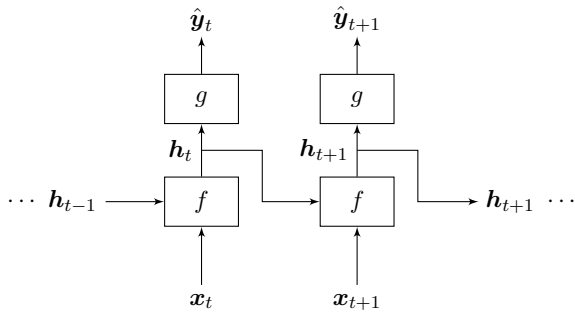

In the conventional mixup setting for feed-forward networks, we $\lambda$-mix randomly chosen samples from the training set (with $\lambda\in\left[0,1\right]$, and usually sampled from a $\mathrm{Beta}\left(\alpha,\alpha\right)$ distribution) and then expose them to the network for learning. Here, by $\lambda$-mixing of two feature-label pairs $\left(\boldsymbol{z},\boldsymbol{y}\right)$ and $\left(\boldsymbol{z}',\boldsymbol{y}'\right)$, we mean the artificial pair 
$\left(\lambda\boldsymbol{z}+\left(1-\lambda\right)\boldsymbol{z}',\lambda\boldsymbol{y}+\left(1-\lambda\right)\boldsymbol{y}'\right)$ \footnote{In this example, both feature $\boldsymbol{z}$ and corresponding one-hot label $\boldsymbol{y}$ are vectors, which are different from the RNN training dataset in this paper where input sample-pairs are sequences of length $T$.}. For the case of an RNN, this procedure becomes complicated since one can think of mixing different timesteps of input sequence pair $\left(\boldsymbol{x},\boldsymbol{y}\right)$ with potentially different mixing coefficients $\lambda$. In fact, the sequence $\boldsymbol{\lambda}=\left\{\lambda_t\right\}$ can be {it {naturally}} considered as a time-dependent stochastic process with preferably $\mathrm{Beta}\left(\alpha,\alpha\right)$-distributed marginals. 

To summarize our contribution, first, we introduce a computationally efficient way of drawing instances of $\left\{\lambda_t\right\}$ with an arbitrary level of temporal correlation in section \ref{sec:mixup-coefficients}. Also, in Section \ref{subsec:cor-mixup-cof}, we investigate the role of temporal correlation among $\lambda_t, t\in\mathcal{T}$ in the performance of models. Moreover, we show that extending manifold mixup of \cite{verma2018manifold} to the realm of RNNs can be done in several ways. In this regard, we propose two possible extensions and give an asymptotic theoretical analysis for them in Section \ref{sec:theory}. A detailed experimental investigation is also presented in Sections \ref{subsec:spec-compres} and \ref{subsec:evaluation}, respectively.


\subsection{Algorithms}
\label{sec:seqMixup}

For each step of training with a mini-batch size of $1$ (extension to larger mini-batch sizes is straightforward), given mixup coefficients $0 \leq \lambda_t \leq 1$ for all $t\in\mathcal{T}$, and assuming two randomly selected sample sequences $\left(\boldsymbol{x}, \boldsymbol{y}\right)$ and $\left(\boldsymbol{x}^{\prime}, \boldsymbol{y}^{\prime}\right)$ from the training dataset $\boldsymbol{D}$, we propose a natural extension of input mixup as follows:
\begin{itemize}
\item
{\bf {Sequence Input Mixup}} algorithm at each time step $t$, replaces $\boldsymbol{x}_t$ by $\boldsymbol{x}_t^{\mathrm{mix}}(\lambda_t)$ and $\boldsymbol{y}_t$ by $\boldsymbol{y}_t^{\mathrm{mix}}(\lambda_t)$ for the two selected samples $\left(\boldsymbol{x},\boldsymbol{y}\right)$ and $\left(\boldsymbol{x}^{\prime}, \boldsymbol{y}^{\prime}\right)$, and for all $t\in\mathcal{T}$. This procedure is illustrated in Figure \ref{dig:im}.
\end{itemize}
\noindent
Also, we present two possible extensions for manifold mixup as: 
\begin{itemize}
\item 
\textbf{Pre-Output Mixup (POM)}, where mixup occurs before the output layer as shown in Figure \ref{dig:pom}, and
\item
\textbf{Through-Time Mixup (TTM)}, where mixup happens at the hidden layers which produce and then propagate the mixed hidden state  as shown in Figure \ref{dig:ttm}.
\end{itemize}

The main difference between POM and TTM is that the latter keeps only one shared hidden representation for both sample pairs, while the former allows each pair to have their own flow historical information. More specifically, POM replaces $g\left(\boldsymbol{h}_t\right)$ by $g\left(\boldsymbol{h}_t^{\mathrm{mix}}\left(\lambda_t\right)\right)$ and $\boldsymbol{y}_t$ by $\boldsymbol{y}_t^{\mathrm{mix}}(\lambda_t)$, while TTM replaces $\boldsymbol{h}_t$ by $\boldsymbol{h}^{\mathrm{mix}}_t(\lambda_t)$ and $\boldsymbol{y}_t$ by $\boldsymbol{y}_t^{\mathrm{mix}}(\lambda_t)$. We have skipped the detailed algorithmic explanation of mixup training in this paper for the sake of readability. An interested reader can find such information in the original papers by \cite{zhang2017mixup} and \cite{verma2018manifold}, respectively.
    
\tikzset{
f/.style = {draw, fill=white, rectangle, minimum height=2em, minimum width=3em},
g/.style = {draw, fill=white, rectangle, minimum height=2em, minimum width=3em},
mix/.style= {draw, fill=white, circle, node distance=1.5cm},
tmp/.style  = {coordinate}
}
    
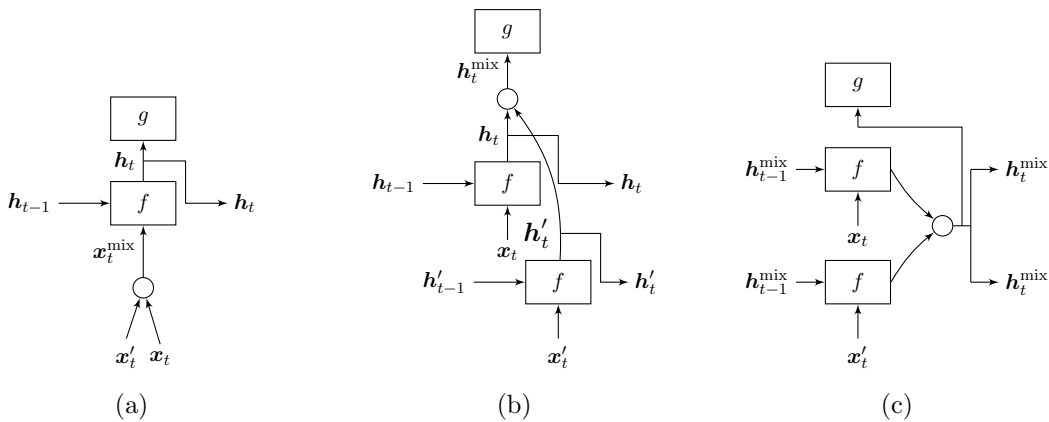
\begin{figure}[t]
    \begin{subfigure}[b]{0.32\textwidth}
     \centering
     \begin{tikzpicture}[auto, node distance=1.5cm,>=latex',scale=0.75, every node/.style={scale=0.75}]
        \node [f, name=f1] (f1) {$f$};
        \node [g, above of=f1] (g1) {$g$};
        \node [mix, below of=f1] (mix1) {};
        \node [tmp, left of=f1, node distance=1.5cm] (tmp1) {};
        \node [tmp, right of=f1, node distance=0.75cm] (tmp2) {};
        \node [tmp, right of=tmp2, node distance=0.75cm] (tmp3) {};
        \draw[->] (mix1) -- node[node distance=0.5cm]{$\boldsymbol{x}^{\mathrm{mix}}_t$} (f1) ;
        \draw[->] (f1) -- node[node distance=0.5cm, name=ht]{$\boldsymbol{h}_t$} (g1);
        \draw [->] ($(mix1)-(-0.3,1cm)$) node[below]{$\boldsymbol{x}_t$} -- (mix1);
        \draw [->] ($(mix1)-(0.3,0.9cm)$) node[below]{$\boldsymbol{x}^{\prime}_t$} -- (mix1);
        \draw [->] (tmp1) node [left]{$\boldsymbol{h}_{t-1}$} -- (f1);
        \draw [->] (ht) -| (tmp2) |-  node[right of=tmp3, node distance=1.05cm]{$\boldsymbol{h}_t$} (tmp3);
    \end{tikzpicture}
    \caption{}
    \label{dig:im}
    \end{subfigure}
    \begin{subfigure}[b]{0.32\textwidth}
    \centering
    \begin{tikzpicture}[auto, node distance=1.5cm,>=latex',scale=0.75, every node/.style={scale=0.75}]
        \node [f, name=f1] (f1) {$f$};
        \node [mix, above of=f1] (mix1) {};
        \node [g, above of=mix1, node distance=1.2cm] (g1) {$g$};
        \node [tmp, left of=f1, node distance=1.5cm] (tmp1) {};
        \node [tmp, right of=f1, node distance=0.9cm] (tmp12) {};
        \node [tmp, right of=tmp12, node distance=1cm] (tmp13) {};
        \node [f, below of=tmp12, node distance=1.75cm] (f2) {$f$};
        \node [tmp, left of=f2, node distance=1.5cm] (tmp2) {};
        \node [tmp, right of=f2, node distance=0.75cm] (tmp22) {};
        \node [tmp, right of=tmp22, node distance=0.5cm] (tmp23) {};
        \draw [->] ($(f1)-(0,1cm)$) node[below]{$\boldsymbol{x}_t$} -- (f1);
        \draw [->] ($(f2)-(0,1cm)$) node[below]{$\boldsymbol{x}^{\prime}_t$} -- (f2);
        \draw [->] (tmp1) node [left]{$\boldsymbol{h}_{t-1}$} -- (f1);
        \draw [->] (tmp2) node [left]{$\boldsymbol{h}^{\prime}_{t-1}$} -- (f2);
        \draw[->] (f1) -- node[node distance=0.5cm, name=ht]{$\boldsymbol{h}_t$} (mix1);
        \path[every node/.style={font=\sffamily\small}]
         (f2) edge [->, bend right=20] node [left=0, pos=0.15, name=hpt] {$\boldsymbol{h}^{\prime}_t$} (mix1);
        \draw [->] (ht) -| (tmp12) |-  node[right of=tmp13, node distance=1.3cm]{$\boldsymbol{h}_t$} (tmp13);
        \draw [->] (hpt) -| (tmp22) |-  node[right of=tmp23, node distance=0.8cm]{$\boldsymbol{h}^{\prime}_t$} (tmp23);
        \draw [->] (mix1) -- node[node distance=0.5cm]{$\boldsymbol{h}^{\mathrm{mix}}_t$} (g1);
    \end{tikzpicture}
    \caption{}
    \label{dig:pom}
    \end{subfigure}
    \begin{subfigure}[b]{0.32\textwidth}
     \centering
     \begin{tikzpicture}[node distance=1.5cm,>=latex',scale=0.75, every node/.style={scale=0.75}]
        \node [f, name=f1] (f1) {$f$};
        \node [f, below of=f1, node distance=2cm] (f2) {$f$};
        \node [tmp, left of=f1, node distance=1.1cm] (tmp1) {};
        \node [tmp, right of=f1, node distance=1.5cm] (tmp12) {};
        \node [tmp, right of=tmp12, node distance=0.5cm] (tmp13) {};
        \node [tmp, right of=tmp13, node distance=0.5cm] (tmp14) {};
        \node [tmp, left of=f2, node distance=1.1cm] (tmp2) {};
        \node [tmp, right of=f2, node distance=1.5cm] (tmp22) {};
        \node [tmp, right of=tmp22, node distance=0.5cm] (tmp23) {};
        \node [tmp, right of=tmp23, node distance=0.5cm] (tmp24) {};
        \node [mix, below of=tmp12, node distance=1cm] (mix1) {};
        \node [tmp, right of=mix1, node distance=0.5cm] (tmpmix){};
        \node [tmp, above of=f1, node distance=.75cm] (tmpg) {$g$};
        \node [g, above of=tmpg, node distance=.75cm] (g1) {$g$};
        \draw [->] ($(f1)-(0,1cm)$) node[below]{$\boldsymbol{x}_t$} -- (f1);
        \draw [->] ($(f2)-(0,1cm)$) node[below]{$\boldsymbol{x}^{\prime}_t$} -- (f2);
        \draw [->] (tmp1) node [left]{$\boldsymbol{h}^{\mathrm{mix}}_{t-1}$} -- (f1);
        \draw [->] (tmp2) node [left]{$\boldsymbol{h}^{\mathrm{mix}}_{t-1}$} -- (f2);
        \draw [->] (f1.east) to [bend right=10] (mix1);
        \draw [->] (f2.east) to [bend left=10] (mix1);
        \draw [-] (mix1) -- node[name=t]{} (tmpmix);
        \draw [-] (tmpmix) -- (tmp13);
        \draw [-] (tmpmix) -- (tmp23);
        \draw [->] (tmp13) -- node[right of=tmp14, node distance=0.75cm]{$\boldsymbol{h}^{\mathrm{mix}}_{t}$} (tmp14);
        \draw [->] (tmp23) -- node[right of=tmp24, node distance=0.75cm]{$\boldsymbol{h}^{\mathrm{mix}}_{t}$} (tmp24);
        \draw [-] ($(t)$) |- (tmpg);
        \draw [->] (tmpg) -- (g1);
    \end{tikzpicture}
    \caption{}
    \label{dig:ttm}
    \end{subfigure}
    \caption{Block diagram of the proposed Sequence Input Mixup \subref{dig:im}, Pre-Output Mixup \subref{dig:pom}, and Through-Time Mixup \subref{dig:ttm}. Circle denote the mixup operation.}
    \label{fig:mixup-dig}
\end{figure}

\begin{remark}
Even though we originally propose sequence mixup for sequence tagging, adopting each of its methods for other similar tasks is straightforward. For example in sequence classification, mixing the labels is equivalent to replacing sequence label $\boldsymbol{y}$ with $\boldsymbol{y}^{\mathrm{mix}}(\bar{\lambda})$, where $\bar{\lambda}$ is the empirical mean of process $\left\{\lambda_t\right\}$.
\end{remark}
    

\subsection{Temporal Dependency Of Mixing Coefficients} 
\label{sec:mixup-coefficients}

In order to choose $\lambda$, \cite{zhang2017mixup} and \cite{verma2018manifold} employed the Beta distribution, $\mathrm{Beta}(\alpha, \alpha)$, for selecting coefficients in non-recurrent neural networks, where hyper-parameter $\alpha$ is usually adjusted for each particular task. Following the same strategy, we have also used the Beta distribution to generate mixup coefficient sequence $\left\{\lambda_t\right\}$. However, the set of  coefficients $\left\{\lambda_t\right\}$ are time-series data and therefore can be correlated through time. In fact, since both input and output sequences have potentially important time-varying dependencies, it is natural to design $\lambda_t$s with some levels of temporal correlation. In section \ref{sec:experiments}, we have experimentally shown that correlation level of mixup coefficients $\left\{\lambda_t\right\}$ has a meaningful impact on the performance of the model. Interestingly, for each particular task, fixing the spectral bandwidth of $\left\{\lambda_t\right\}$ to a corresponding optimal value across the time $\mathcal{T}$ reduces the loss of the base-line model.

In order to generate the mixup coefficients, we create a non-stationary Markov process with varying correlation levels as follows:
\begin{align*}
\lambda_{1} &\sim \mathrm{Beta}\left(\alpha, \alpha\right) \\    
\lambda_{t} | \lambda_{t-1} &\sim \mathrm{Beta}\left(\alpha_t, \beta_t\right) \quad \forall t \geq 2,
\end{align*}
where $\alpha_t=\alpha_t\left(\lambda_{t-1}\right)$ and $\beta_t=\beta_t\left(\lambda_t\right)$ are computed by solving the following set of equations:
\begin{align}
\left\{\begin{array}{rl}
\mathbb{E}\left[\lambda_{t} | \lambda_{t-1}\right] &= \quad\quad~\frac{\alpha_t}{\alpha_t+\beta_t}\quad\quad~=\rho \mathbb{E}\left[\lambda_1 \right] + \left( 1-\rho \right) \lambda_{t-1}
\nonumber
\\[1mm]
\mathrm{Var}\left(\lambda_t | \lambda_{t-1} \right) &= \frac{\alpha_t\beta_t}{\left(\alpha_t+\beta_t\right)^2\left(\alpha_t+\beta_t+1\right)} = \rho^2 \mathrm{Var}\left( \lambda_1  \right)
\end{array}
\right.
\end{align}
There are closed-form formulas for efficiently computing $\left(\alpha_t,\beta_t\right)$s for each time-step. Here, user-defined hyper-parameter $ 0 \leq \rho \leq 1$ controls the level of dependency through time. Specifically, $\rho$ smooths the trajectory of coefficients by forcing $\lambda_{t}$ to be close to  $\lambda_{t-1}$ through fixing its conditional mean near the $\lambda_{t-1}$ and reducing its conditional variance.


\subsection{Asymptotic Theoretical Analysis}
\label{sec:theory}

In this part, we present the main theoretical results regarding the asymptotic analysis of sequence mixup, and Pre-Output Mixup (POM) technique in particular. Here, we attempt to give a more intuitive insight regarding our theoretical findings while mathematical details, notations and formal definitions are explained in Appendix \ref{sec:app:theory}. Roughly speaking, the term ``asymptotic" refers to the following two assumed properties:
\begin{itemize}
\item 
Hidden layer widths (or depths, but not necessarily both) of the neural networks corresponding to functions $f$ and $g$ in Figure \ref{fig:mixup-dig} become asymptotically large.
\item
The learning rate which governs the training stage becomes infinitesimally small, which forces the number of training iterations to be increasingly large. This property makes it safe to assume that all possible pairs of samples in the batch have been mixed several times during the training stage.
\end{itemize}

As already mentioned in Section \ref{sec:intro}, let us denote the number of output neurons for function $f$ as $H\in\mathbb{N}$. Also, it should be reminded that the number of classes in the classification problem has been denoted by $C\ge2$. Then, we initially prove the following property for both POM and TTM:

\begin{thm}[Over-Regularization]
\label{thm:overreg}
For any given RNN architecture and classification task, assume we have $H<C-1$. Then, for an infinitesimally small learning rate $\eta$, any training dataset size $n$, and regardless of the vertical or horizontal sizes of neural networks corresponding to functions $f$ and $g$, training error of both POM and TTM cannot strictly become, or asymptotically approach toward, zero.
\end{thm}

Proof of Theorem \ref{thm:overreg} is discussed in Appendix \ref{sec:app:theory}. Theorem \ref{thm:overreg} states that when $H$ is chosen to be smaller than $C-1$, with $C$ being the number of classes in the problem, one just over-regularizes the RNN via POM or TTM. In other words, the RNN cannot be trained to zero-force the training error regardless of the complexity of its built-in neural nets and thus suffers from a non-zero bias error \footnote{Note that due to Universal Approximation Theorem \citep{cybenko1989approximation}, non-regularized and asymptotically large neural nets always completely overfit to a finite-size training dataset, i.e., training error always become or asymptotically approach towards zero.}. The following two theorems only hold for POM, however, similar arguments might hold for TTM as well.

\begin{thm}
\label{thm:POMmemory}
For any given RNN architecture and classification task, assume we have $H<2C-1$. Then, for any training dataset size $n$, an infinitesimally small learning rate $\eta$, and asymptotically large vertical and/or horizontal sizes of the neural networks corresponding to $f$ and $g$, the following argument holds for the asymptotic solution of POM, denoted by $f^*:\mathbb{R}^{H}\times \mathbb{R}^d\rightarrow\mathbb{R}^H$~: 

$f^*$ acts as an almost memory-less unit, i.e., there exists function $\Tilde{f}:\mathbb{R}^d\times\mathcal{Y}\rightarrow\mathbb{R}^H$ such that for all $t\in\mathcal{T}\char`\\\left\{1\right\}$:
$$
f^*\left(\boldsymbol{x}_t;\boldsymbol{h}_{t-1}\right)=\Tilde{f}\left(\boldsymbol{x}_t;\boldsymbol{y}_{t-1}\right).
$$
\end{thm}

Theorem \ref{thm:POMmemory} roughly states that in order to enable POM solution $f^*$ to use the information of previous time-steps (through $\boldsymbol{h}_t$), one has to make sure that $H$ is at least almost twice the number of distinct classes $C$. Otherwise, $f^*$ only makes use of the previous label $\boldsymbol{y}_{t-1}$ and current feature $\boldsymbol{x}_t$ to estimate $\boldsymbol{y}_t$, which means the state vector $\boldsymbol{h}_t$ that carries the information of all previous time-steps would be ignored. The final theorem which is stated below is the most important property of POM (and probably TTM), which describes their ability to linearly separate different classes in the representation space.

\begin{thm}[Spectral Compression Property]
For any given RNN architecture, classification task, and training dataset size $n$, assume we have $H\ge2C-1$. Also, assume POM is used for regularization with an infinitesimally small learning rate $\eta$, asymptotically large vertical and/or horizontal sizes of neural nets corresponding to $f$ and $g$, and an arbitrary mini-batch size for training. Also, let us denote the learned RNN functions by $f^*$ and $g^*$, respectively. Then, the output-generating function $g^*$ is (at least locally) linear, i.e., there exists $\boldsymbol{W}_g\in\mathbb{R}^{H\times C}$ and $\boldsymbol{b}_g\in\mathbb{R}^{C}$ such that
$$
g^*\left(\boldsymbol{h}\right)=\boldsymbol{W}^T_g\boldsymbol{h} + \boldsymbol{b}_g,\quad\forall\boldsymbol{h}\in\mathrm{Conv}\left(\left\{\boldsymbol{h}^{\left(i\right)}_{t}\right\}_{i=1,\ldots,n}^{t=1,\ldots,T}\right),
$$
where $\mathrm{Conv}\left(\cdot\right)$ denotes the convex hull of a set of points. Also, $f^*$ divides the representation space $\mathbb{R}^H$ into a set of $C$ orthogonal affine subspaces $\left\{\mathcal{S}_1,\ldots,\mathcal{S}_C\right\}$, such that for all $i=1,\ldots,C$, we have:
$
\left\{f^*\left(\boldsymbol{x}_t;\boldsymbol{h}_{t-1}\right)\big\vert~\boldsymbol{y}_t=\boldsymbol{e}_i
\right\}\subseteq\mathcal{S}_i,
$
where $\boldsymbol{e}_i$ represents a one-hot vector with its $i$th component being $1$ and the rest being $0$. Also, the following equality holds w.r.t. dimensions of affine subspaces $\mathcal{S}_1,\ldots,\mathcal{S}_C$:
$$
\sum_{i=1}^{C}\mathrm{dim}\left(\mathcal{S}_i\right)=H-C+1.
$$
\end{thm}

Spectral compression property is the main essence of hidden-state-based mixup strategies, i.e., manifold mixup for feed-forward nets and the proposed Pre-Output Mixup for RNNs. It shows that the mixup regularizer, when applied to the hidden states of neural nets, forces the core network to map data samples with different labels into distinct orthogonal affine subspaces of the representation space. Also, it dictates that the output-generating network mimics a simple linear unit. This property, which is partially validated through our experiments in Section \ref{sec:experiments} (even for non-asymptotic neural architectures), means that the spectral power distribution of hidden-state vectors for samples with the same label becomes more compact compared to non-regularized cases.

    
\section{Experiments}\label{sec:experiments}
In this section, three sets of experiments have been provided: In Section \ref{subsec:cor-mixup-cof}, we have studied the performance of sequence mixup as the correlation between mixup coefficients varies in time. In Section \ref{subsec:spec-compres}, spectral compression property of POM and TMM have been studied. Finally, in Section \ref{subsec:evaluation}, we have compared the performance of sequence mixup with standard training. All experiments have been conducted on the named entity recognition task over CoNLL-2003 data \citep{sang2003introduction}.

\subsection{Correlation of Mixup Coefficients}
\label{subsec:cor-mixup-cof}

We define the baseline model as a model consisting of an embedding layer initialized with weights of Glove embedding \citep{pennington2014glove} followed by a single layer recurrent network with hidden size of $H=256$, and finally a layer to map hidden states to class scores.

\begin{figure*}[t]
	\centering
	\subfloat[]{\includegraphics[width=0.49\linewidth]{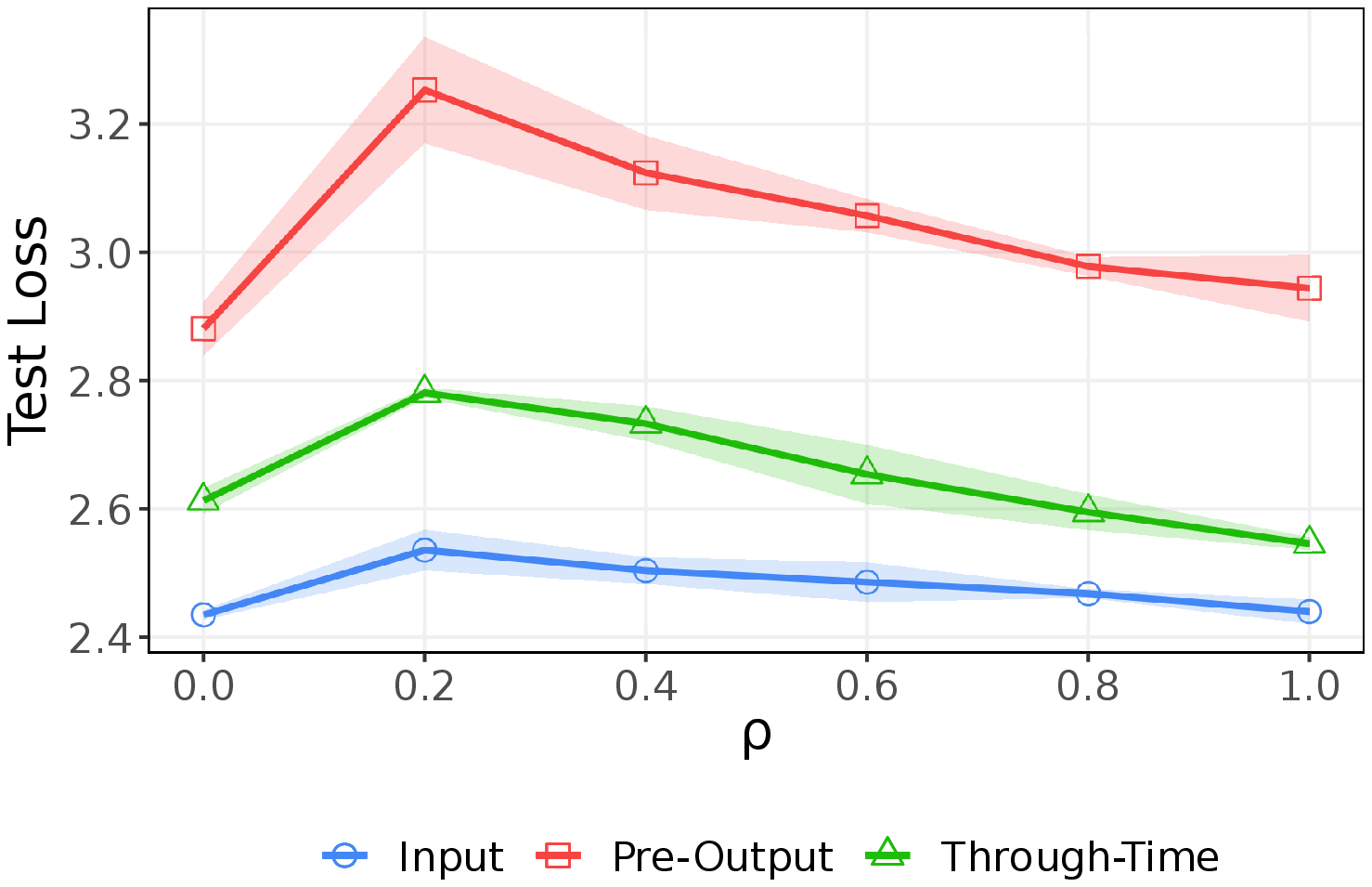}  \label{fig:rho-loss}}
	\subfloat[]{\includegraphics[width=0.49\linewidth]{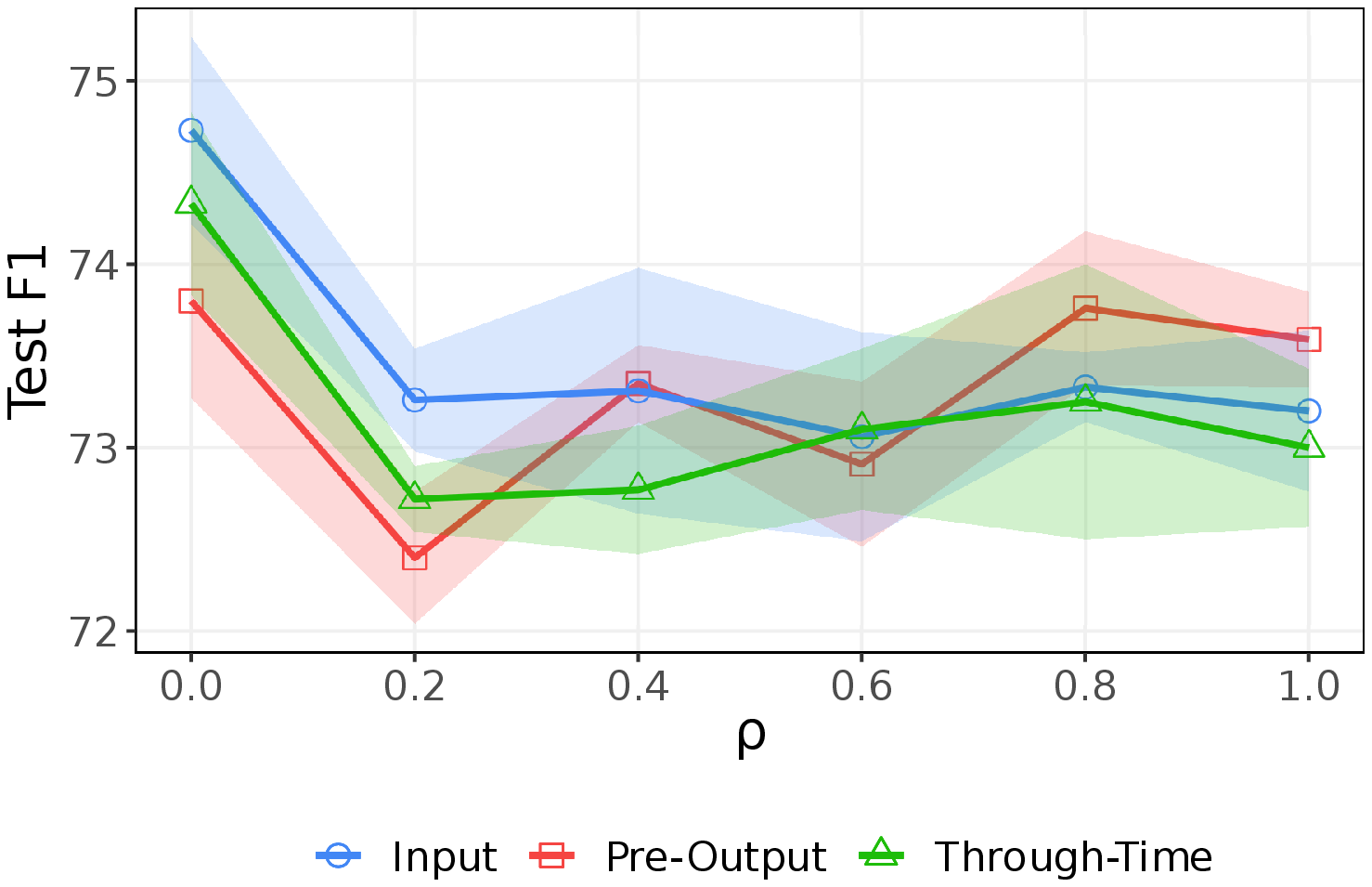}  \label{fig:rho-f1}}
	\caption{Relationship between $\rho$ and \subref{fig:rho-loss} the model's loss and \subref{fig:rho-f1} F-1 score on the test data when trained with corresponding methods.}
	\label{fig:rho}
\end{figure*}

We have trained the baseline model with LSTM cell, $50$ epochs using stochastic gradient descent with an initial learning rate of $0.1$ (while halving it after each $10$ epochs). Also, cross-entropy is chosen as the measure of loss. Sequence mixup training with mixup coefficients $\left\{\lambda_t\right\}$ generated from the Markov process of Section \ref{sec:mixup-coefficients} (with $\alpha=0.5$ and varying $\rho$) have been separately utilized for training. Figure \ref{fig:rho} illustrates the model's F-1 score and loss on the test data as a function of $\rho$ for Sequence Input Mixup, Pre-Output Mixup and Through-Time Mixup, respectively. As it is evident, the choice of $\rho=0$, which is equivalent to setting all $\lambda_t$ identical, has maximum test F-1 on all methods. Also, choosing a particular intermediate value for $\rho$ has the worst effect on test loss.


\subsection{Spectral Compression}
\label{subsec:spec-compres}

Based on the theoretical results in Section \ref{sec:theory}, POM (and possibly TTM) divide the hidden state of an RNN to orthogonal affine subspaces in the $H$-dimensional representation space, where each subspace associates to a particular class index. This behavior will compress hidden states of samples within the same class into a lower-dimensional space. We have tried Singular Value Decomposition (SVD) to capture this effect by analyzing the spectral power compactness of the corresponding singular values.

\begin{figure*}[!b]
	\centering
	\subfloat[]{\includegraphics[width=0.49\linewidth]{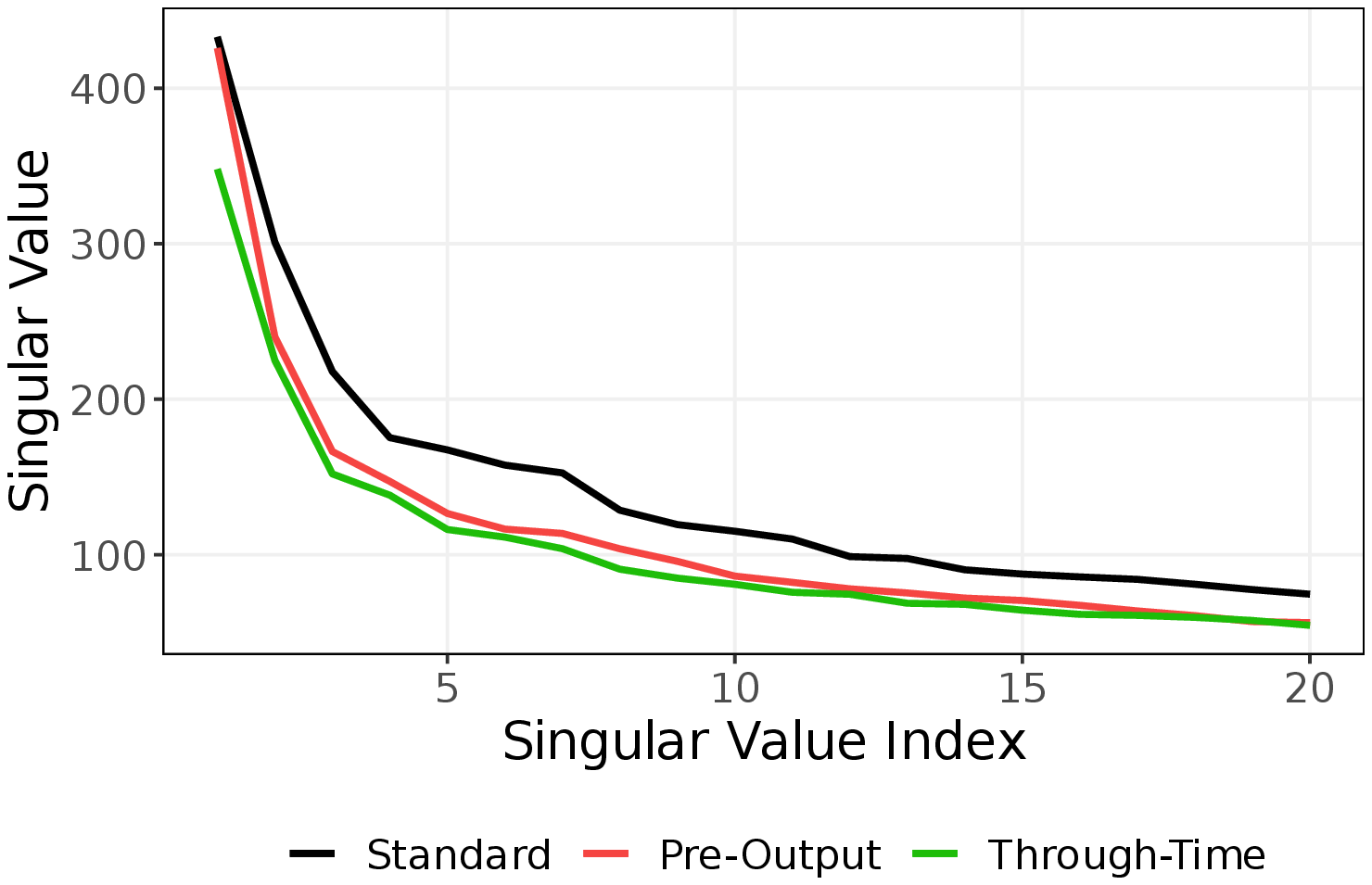}  \label{fig:sv-O}}
	\subfloat[]{\includegraphics[width=0.49\linewidth]{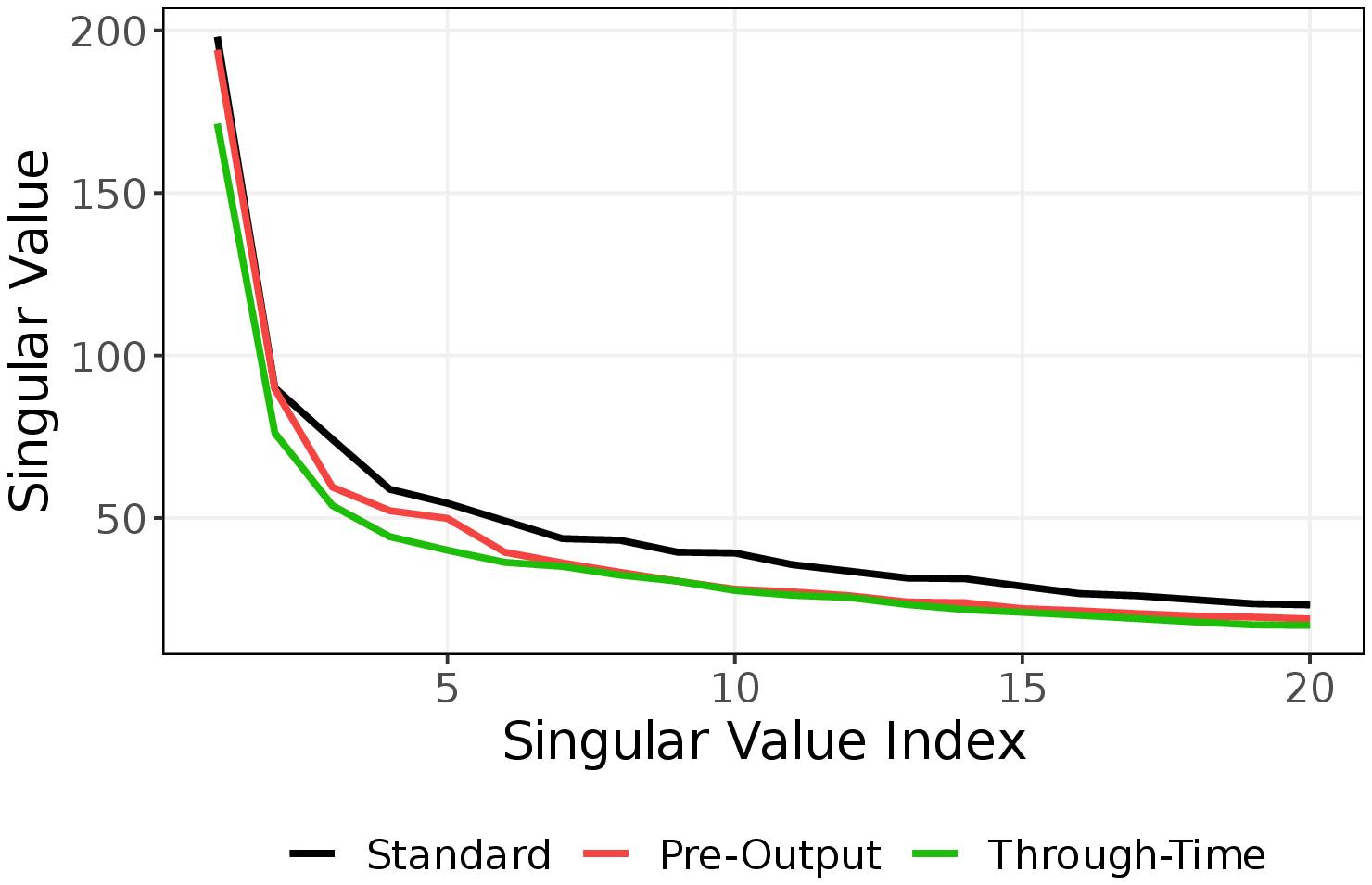}  \label{fig:sv-PER}}
	\caption{The ordered $20$ largest singular values corresponding to the hidden state vectors ($\boldsymbol{h_t}$) of the same label, for two random and different classes.}
	\label{fig:sv}
\end{figure*}

Figure \ref{fig:sv} plots the $20$ largest singular values obtained from hidden state vectors of training data which correspond to the same label, for two randomly selected and distinct labels. The baseline model with LSTM cell and $\alpha=1$ is employed and subsequently trained with regular training, POM, and TTM, respectively. As demonstrated in the figure, more compact singular value distributions for POM and TTM suggest that hidden states for such methods lie on a compact and lower-dimensional subspace compared to the case of standard training. This observation is in agreement with the Spectral Compression Property, which is proved for asymptotic network architectures in Section \ref{sec:theory}.

\subsection{Evaluation}
\label{subsec:evaluation}
    
Table \ref{tbl:alpha-cmp} shows the F-1 scores of the baseline model on the test data with various cell types when trained with and without Sequence Mixup for different values of $\alpha$. We have set $\rho=0$ for generating mixup coefficients, as suggested by the previous result. Details of the experiments are similar to that of Section \ref{subsec:cor-mixup-cof}.

\setlength{\arrayrulewidth}{0.75pt}
\begin{table}[!b]
    \centering
    \begin{tabular}{c|c|cccc}
    \hline
    cell & Standard & alpha & Sequence Input Mixup &  POM & TTM\\
    \hline
    \hline
    \multirow{5}{*}{RNN} 
       & & $0.1$ & $72.25 \pm 0.45$ & $71.51 \pm 0.24$ & $72.14 \pm 0.38$\\
       & & $0.2$ & $72.40 \pm 0.23$ & $71.75 \pm 0.61$ & $72.30 \pm 0.33$ \\
       & $70.72 \pm 0.21$ & $0.5$ & $71.53 \pm 0.42$ & $72.32 \pm 0.55$ &  $72.57	\pm 0.22$ \\
       & & $1$   & $72.58 \pm 0.63$ & $71.54 \pm 0.43$ & $72.15 \pm 0.31$\\
       & & $2$   & $72.48 \pm 0.32$ & $70.68 \pm 0.35$ & $71.96 \pm 0.34$\\
    \hline
    \multirow{5}{*}{GRU} 
       & & $0.1$ & $73.55 \pm 0.42$ & $73.48 \pm 0.17$ & $73.37 \pm 0.45$\\
       & & $0.2$ & $73.78 \pm 0.34$ & $73.36 \pm 0.29$ & $73.86 \pm 0.45$\\
       & $72.17 \pm 0.30$ & $0.5$ & $73.85 \pm 0.51$ & $73.05 \pm 0.33$  & $73.74 \pm 0.58$ \\
       & & $1$   & $74.00 \pm 0.40$ & $72.90 \pm 0.54$  & $73.33 \pm 0.38$\\
       & & $2$   & $73.32 \pm 0.33$ & $72.93 \pm 0.32$  & $73.33 \pm 0.61$\\
    \hline
    \multirow{5}{*}{LSTM} 
       & & $0.1$ & $74.35 \pm 0.30$ & $73.67 \pm 0.38$  & $73.84 \pm 0.17$\\
       & & $0.2$ & $74.35 \pm 0.31$ & $73.78 \pm 0.61$  & $74.34 \pm 0.70$\\
       & $73.23 \pm 0.31$ & $0.5$ & $74.73 \pm 0.51$ & $73.80 \pm 0.53$ & $74.33 \pm 0.50$\\
       & & $1$   & $74.34 \pm 0.54$ & $73.94 \pm 0.22$  & $73.96 \pm 0.32$\\
       & & $2$   & $74.28 \pm 0.33$ & $73.77 \pm 0.30$  & $73.39 \pm 0.65$\\
    \end{tabular}
    \caption{F-1 score of the baseline model with different cells and various values of $\alpha$ on CoNLL-2003 Named Entity Recognition.}
    \label{tbl:alpha-cmp}
\end{table}
    
In Table \ref{tbl:biLSTM-CRF}, we have trained biLSTM-CRF model \citep{huang2015bidirectional} with combination of contextual word embeddings \citep{akbik2018contextual} and Glove embeddings \citep{pennington2014glove}. The model specification is the same as \cite{huang2015bidirectional} for CoNLL-2003 on named entity recognition task. Specifically, there is an embedding layer initialized with contextual word embeddings and Glove, followed by a single-layer bidirectional LSTM with hidden size of $256$, and finally a linear layer to generate the class scores. Similarly, for training, we followed the \cite{akbik2018contextual} and trained the model using Vanilla stochastic gradient descent, clipping gradients at $5$ for $250$ epochs with batch size of $32$. For scheduling the learning rate, we halve it if training loss doesn't improve for $6$ consecutive epochs. When training the model without mixup, we use locked dropout \citep{merity2017regularizing} and word-level dropout as used by \cite{akbik2018contextual}. However, while using Sequence Mixup, we only utilize locked dropout to avoid the over-regularization problem. We perform model selection over $\alpha \in \left\{0.1, 0.2, 0.5, 0.7, 1\right\}$, choosing the model with the minimum F-1 score on the validation set. Mixup coefficients then set to a random sample from $\mathrm{Beta}\left(\alpha, \alpha\right)$ across the sentence -- same as setting $\rho=0$ in the previous part, and the model trains with both training and validation set. Experiments are then repeated $5$ times to obtain the mean and standard deviation of the F-1 score and loss, respectively.
    
Denoting target sequence of length $T$ by $\boldsymbol{y}$, predicted class scores by $\hat{\boldsymbol{y}}$ and transition matrix of the CRF model by $\boldsymbol{A}$, biLSTM-CRF model of \cite{huang2015bidirectional} scores each sequence by\footnote{For the ease of notations, here, $y_t$ denotes the class index which corresponds to the already-defined one-hot vector $\boldsymbol{y}_t$.}
$$
\sum_{t=1}^T{\hat{y}_{t, y_t}} + \sum_{t=1}^{T-1}{A_{y_t, y_{t+1}}},
$$
and computes the loss function as cross-entropy between one-hot label of the whole sequence among all possible sequences and the distribution of sequence scores. Applying Sequence Mixup methods, converts each sequence score to 
\begin{align*}
\sum_{t=1}^T{\lambda_t \hat{y}_{t, y_t} + (1-\lambda_t)\hat{y}_{t, y^{\prime}_t}}
+ 
\sum_{t=1}^{T-1}{\left(\frac{\lambda_t 
+ 
\lambda_{t+1}}{2}\right)A_{y_t, y_{t+1}} + \left(1-\frac{\lambda_t + \lambda_{t+1}}{2}\right)A_{y^{\prime}_t, y^{\prime}_{t+1}}},
\end{align*}
for two samples $\left(\boldsymbol{x}, \boldsymbol{y}\right)$ and $\left(\boldsymbol{x}^{\prime}, \boldsymbol{y}^{\prime}\right)$.
    
\begin{table}[!b]
    \centering
    \begin{tabular}{lcc}
    \toprule
        biLSTM-CRF & F-1 & Test NLL\\
    \midrule
    \midrule
        Standard & $93.22 \pm 0.09$ & $1.07 \pm 0.03$
    \\ \midrule
        Sequence Input Mixup ($\alpha=0.1$) & $93.29 \pm 0.09$ & $\mathbf{0.59 \pm 0.01}$
    \\ \midrule
        Pre-Output Mixup ($\alpha=0.7$) & $\mathbf{93.30 \pm 0.09} $ & $0.72 \pm 0.01$
    \\ \midrule
        Through-Time Mixup ($\alpha=0.1$) & $93.17 \pm 0.06$ & $0.67 \pm 0.004$
    \end{tabular}
    \caption{Evaluation of biLSTM-CRF model using conventional and Seuquence Mixup training methods on CoNLL-2003 NER task.}
    \label{tbl:biLSTM-CRF}
\end{table}

According to Table \ref{tbl:biLSTM-CRF}, TTM's F-1 on the test data has gotten worse than standard training. This may be due to the fact that TTM is a much stronger regularizer than the other two, which them means it does not need locked dropout as an extra regularization. Interestingly, the loss of Sequence Mixup methods is much smaller than regular training, which consolidates the claim that Sequence Mixup renders a decision boundary with less certainty about difficult-to-classify instances.


\section{Conclusion}
\label{sec:conclusion}

We introduce Sequence Mixup, a set of regularization and data augmentation techniques for RNNs. Our work can thought as extending both input mixup \citep{zhang2017mixup} and manifold mixup \citep{verma2018manifold}, which are originally porposed for feed-forward neural nets. For the case of manifold mixup, we propose two distinct variants called Pre-Output and Throgh-Time Mixup, respectively. An asymptotic theoretical analysis reveals that Pre-Output Mixup imposes (at least) a locally linear behavior on the network's output generating section. In a classification task, this property leads to partitioning of the hidden representation space into a set of orthogonal affine subspaces, each of which corresponds to a unique class. Experimental results showed improvement on the loss and F-1 scores of both 1) a baseline and 2) state-of-the-art model on CoNLL-2003 NER task. We have studied the correlation of mixup coefficients through consecutive time-steps, and found out that using identical coefficients achieves better loss and F-1 on the NER task. However, at the same time, we conjecture that optimal correlation values for mixup coefficients across time may vary from task to task and thus requires experimental exploration to be adjusted. Lastly, the considerable reduction in the test loss achieved by sequence mixup methods (Section \ref{subsec:evaluation}) implies that employing sequence mixup methods for language models may lead to a substantial improvement on the test perplexity.

\bibliography{refs.bib}
\bibliographystyle{chicago}

\appendix
\section{Asymptotic Theoretical Analysis: Details}
\label{sec:app:theory}

In this section, we theoretically analyze the asymptotic aspects of some of our regularization techniques during the training stage. The term asymptotic here implies that our analysis only considers cases where the width of the neural networks used in the architecture of RNNs, and also the number of SGD iterations in the training stage go to infinity. Non-asymptotic analyzes have also been done w.r,t. neural networks, however, their depth, scope and achievements have been considerably limited due to the inherent technical difficulty of the problem. Consideration of asymptotically wide neural nets (when the number of neurons in the hidden layer goes to infinity) allow us to take advantage of the Universal Approximation Theorem (UAT) \citep{cybenko1989approximation} in order to facilitate the behaviour and performance analysis of neural networks.

First, let us introduce a number of necessary notations that we use in this section. We show the empirical distribution of data samples $\boldsymbol{D}=\left\{\left(\boldsymbol{x}_1,\boldsymbol{y}_1\right),\ldots,\left(\boldsymbol{x}_n,\boldsymbol{y}_n\right)\right\}$ by 
\begin{equation}
\hat{\mathbb{P}}_{\boldsymbol{D}}\triangleq
\sum_{i=1}^{n}\delta_{\boldsymbol{Z}_i},    
\end{equation}
where $\boldsymbol{Z}_i$ denotes the tuple $\left(\boldsymbol{x}_i,\boldsymbol{y}_i\right)$ showing the $i$th training sample consisting of feature vector $\boldsymbol{x}_i$ and one-hot label vector $\boldsymbol{y}_i\in\left\{0,1\right\}^C$. Here, $C$ denotes the number of classes in our supervised learning problem. Let $\mathrm{NN}_{q,r}$ represent the hypothesis set corresponding to neural network classifiers with $C$ output neurons, which have at least $q$ hidden layers, and at least $r$ neurons per each hidden layer. Throughout this section, we assume $f\in\mathrm{NN}_{q,r}$ with $q\ge1$.

Training with Input or Manifold Mixup cannot be readily formulated into a manageable analytic format unless one assumes very small gradient descent steps and a consequently large number of iterations. Let us assume stochastic gradient descent (SGD) with a mini-batch size of $1$ is used to optimize the loss function associated to any of the Input and/or Manifold Mixup regularization techniques described in Section \ref{sec:main}. Also, let $N_{\mathrm{iter}}$ denote the number of SGD iterations during the training stage. In this section, we are particularly interested in an asymptotic regime where both $N_{\mathrm{iter}}$ and $r$ go to infinity. In other words, we set out to investigate the properties of the solution neural networks $f^*$ and $g^*$ defined as
\begin{align}
\left(f^*,g^*\right)\triangleq
\lim_{r\rightarrow\infty}
\argmin_{\left(f,g\right)\in\mathrm{NN}_{q,r}}~\left(\lim_{N_{\mathrm{iter}}\rightarrow\infty}\hat{\ell}_{\mathrm{emp}}\right).
\end{align}
We show that regardless of the size of the training dataset $n$, or the inherent hardness of the learning task, under ceration conditions there exist $f^*$ and $g^*$ which give a zero value for the empirical loss function $\hat{\ell}_{\mathrm{emp}}$.


\subsection{Asymptotic Theory of Pre-Output Mixup (POM)}
In this part, we analyze Pre-Output Mixup (POM) regularization technique. Assuming an infinitely large number of training iterations, the following equality holds almost surely according to the law of large numbers:
\begin{align*}
\lim_{N_{\mathrm{iter}}\rightarrow\infty}\hat{\ell}_{\mathrm{emp}}\stackrel{a.s}{=}
\mathbb{E}_{\left\{\lambda_t,\boldsymbol{Z}_t,\boldsymbol{Z}'_t\right\}}\left\{
\sum_{t\in\mathcal{T}}
\ell\bigg[
g\left(
\lambda_t\boldsymbol{h_t}+\left(1-\lambda_t\right)\boldsymbol{h}'_t
\right),~
\lambda_ty_t + \left(1-\lambda_t\right)y'_t
\bigg]
\right\},
\end{align*}
where $\boldsymbol{Z}_t$ and $\boldsymbol{Z}'_t$ are training sequences randomly and independently chosen from the empirical data distribution $\hat{\mathbb{P}}_{\boldsymbol{D}}$, $\boldsymbol{h}_t$ and $\boldsymbol{h}'_t$ are temporal state vectors $f\left(\boldsymbol{x}_t,\boldsymbol{h}_{t-1}\right)$ and $f\left(\boldsymbol{x}'_t,\boldsymbol{h}'_{t-1}\right)$, respectively, and
$\left\{\lambda_t\right\}$ is an arbitrary Ergodic stochastic process with marginal Beta distributions. The process $\left\{\lambda_t\right\}$ is assumed to be independent from the random choice of training sequence pairs $\boldsymbol{Z}_t$ and $\boldsymbol{Z}'_t$.

Based on the independence assumption between stochastic processes $\left\{\lambda_t\right\}$ and $\left\{\left(\boldsymbol{Z}_t,\boldsymbol{Z}'_t\right)\right\}$, the empirical error for asymptotic regime of $N_{\mathrm{iter}}\rightarrow\infty$ can be rewritten and subsequently simplified as follows:
\begin{align}
\label{eq:indepLoss1}
&\mathbb{E}_{\left\{\lambda_t,\boldsymbol{Z}_t,\boldsymbol{Z}'_t\right\}}\left\{
\sum_{t\in\mathcal{T}}
\ell\bigg[
g\left(
\boldsymbol{h}^{\mathrm{mix}}_t\left(\lambda_t\right)
\right),~
y^{\mathrm{mix}}_t\left(\lambda_t\right)
\bigg]
\right\}
=
\\
&\hspace*{6mm}\frac{1}{n^2}\sum_{i,j=1}^{n}
\mathbb{E}_{\left\{\lambda_t\right\}}\left\{
\sum_{t\in\mathcal{T}}
\ell\bigg[
g\left(
\lambda_t \boldsymbol{h}_{t,i} + \left(1-\lambda_t\right)\boldsymbol{h}_{t,j}
\right),~
\lambda_t \boldsymbol{y}_{t,i} + \left(1-\lambda_t\right)\boldsymbol{y}_{t,j}
\bigg]
\right\},
\nonumber
\end{align}
where $\boldsymbol{A}_{t,i}$ for a finite set of vector sequences $\left\{\boldsymbol{A}_t\right\}$ means the $t$th time-instance of the $i$th member in the set. Equation \eqref{eq:indepLoss1} implies that in order to get a zero training error, one must have the following equality for all $t\in\mathcal{T}$ and $i\in\left[n\right]$:
\begin{align}
\int \ell\bigg[
g^*\left(
\lambda_t \boldsymbol{h}_{t,i} + \left(1-\lambda_t\right)\boldsymbol{h}_{t,j}
\right),~
\lambda_t \boldsymbol{y}_{t,i} + \left(1-\lambda_t\right)\boldsymbol{y}_{t,j}
\bigg] \mathrm{d}\mathbb{P}\left(\lambda_t\vert\boldsymbol{\lambda}_{\mathrm{prev}}\right)=0,
\label{eq:necSufCondEll}
\end{align}
where $\boldsymbol{\lambda}_{\mathrm{prev}}$ indicates the set of all $\lambda_{t'}$ values with $t'<t$ and $\mathbb{P}\left(\lambda_t\vert\boldsymbol{\lambda}_{\mathrm{prev}}\right)$ denotes the conditional probability measure of stochastic process $\left\{\lambda_t\right\}$. Recall that $\left\{\lambda_t\right\}$ has been assumed to be an Ergodic real-valued stochastic process, which makes \eqref{eq:necSufCondEll} to hold only when
\begin{align}
\label{eq:1stCond}
&g^*\left(\boldsymbol{h}_{t,i}\right)=\boldsymbol{y}_{t,i},
\\
\label{eq:2ndCond}
\mathrm{and}\quad&
g^*\left(
\lambda \boldsymbol{h}_{t,i} + \left(1-\lambda\right)\boldsymbol{h}_{t,j}
\right) =
\lambda \boldsymbol{y}_{t,i} + \left(1-\lambda\right)\boldsymbol{y}_{t,j},
\\[1mm]
\mathrm{for all}\quad&
t\in\mathcal{T}, i\in\left[n\right]~\mathrm{and}~ \forall\lambda\in\left[0,1\right].
\nonumber
\end{align}
The condition \eqref{eq:2ndCond} dictates a locally linear behaviour for the presumed optimal neural network $g^*$. In other words, $g^*$ is supposed to act as a linear function between any two pairs of points $\left\{\boldsymbol{h}_{t,i}\vert~i\in\left[n\right],~t\in\mathcal{T}\right\}$. Obviously, the local linear behaviour is satisfied if we just let $g^*$ be a linear function throughout its input space. In fact, an important implication of the above analytic statement regarding $g^*$ is that adding extra hidden layers for the output-generating neural net $g$ in order to make it more flexible for imitating nonlinear relations is totally redundant. This it due to the fact that the whole network still tries to imitate a linear function which can be simply attained with zero hidden layers.

By assuming the global linearity property for $g$, we have the following relation for the optimal function $g^*:\mathbb{R}^{H}\rightarrow\mathbb{R}^{C}$, with $H$ denoting the number of neurons for the hidden state vector $\boldsymbol{h}\in\mathbb{R}^H$:
\begin{align}
g^*\left(\boldsymbol{h}\right)=\boldsymbol{W}_g^T\boldsymbol{h}+b_g,\quad\forall\boldsymbol{h}\in\mathbb{R}^H,
\end{align}
where $\boldsymbol{W}_g$ is an $H\times C$ dimensional matrix and $b_g\in\mathbb{R}$. The following statements still hold even if we only consider the proved local linearity property of $g^*$. However, we omit further analytical details for the sake of simplicity and stick to the global linearity assumption. On the other hand, for all $t\in\mathcal{T},~i\in\left[n\right]$ we must have
\begin{align}
\left\{
\begin{array}{l}
\boldsymbol{h}_{t,i}=f^*\left(\boldsymbol{x}_{t,i}~,~\boldsymbol{h}_{t-1,i}\right) 
\\
\boldsymbol{y}_{t,i}=\boldsymbol{W}^T_g\boldsymbol{h}_{t,i}+b_g
\end{array}
\right.
\end{align}
The equalities above almost surely hold under mild conditions which are derived in the remainder of this section. The fundamental reason for this phenomenon is that we have already assumed $r\rightarrow\infty$; Then, based on the universal approximation theorem, the function set  $\mathrm{NN}_{q,r}$ with $q\ge1$ becomes dense in the space of functions $\left\{f:\mathbb{R}^{d+H}\rightarrow\mathbb{R}^H\right\}$, with $d$ denoting the dimension of input feature vectors. In other words, $f^*$ is simply assumed to be able to approximate any function with an arbitrarily small error bound which solely depends on the network's width $r$ and goes to zero as $r$ is being increased.

Note that one-hot label vectors $\boldsymbol{y}_{t,i}$ are finite. In fact, there are only $C$ possible distinct label vectors $\boldsymbol{e}_1,\ldots,\boldsymbol{e}_C$, where $\boldsymbol{e}_i$ represents a one-hot $C$-dimensional vector where the $i$th component is $1$ and the rest are $0$. In other words, the linear function $g^*$ has been assumed to map the set of $H$-dimensional real-valued vectors $\left\{\boldsymbol{h}_{t,i}\right\}$ to the set of linearly independent vectors $\left\{\boldsymbol{e}_1,\ldots,\boldsymbol{e}_C\right\}$.

Let $\boldsymbol{\mathcal{H}}$ be the $H\times \left(n\left\vert\mathcal{T}\right\vert\right)$ matrix whose columns are members of $\left\{\boldsymbol{h}_{t,i}\right\}$. Also, let $\boldsymbol{\mathcal{Y}}$ be a $C\times\left(n\left\vert\mathcal{T}\right\vert\right)$ matrix, which is similarly attained from $\left\{\boldsymbol{y}_{t,i}\right\}$. Then, we have the following relation:
\begin{equation}
\boldsymbol{W}^T_g\boldsymbol{\mathcal{H}} = \boldsymbol{\mathcal{Y}}-b_g\boldsymbol{1}_{C\times\left(n\left\vert\mathcal{T}\right\vert\right)}.
\label{eq:matEquality}
\end{equation}
The above equality has a number of interesting implications for the optimal neural network $f^*$:
\begin{itemize}
\item 
First, it should be noted that the r.h.s. of \eqref{eq:matEquality} is at least rank $C-1$ (assuming we have at least one observation from each class in the training dataset $\boldsymbol{D}$), while the l.h.s. is at most rank $H$. Then, in order to get a zero training error for Pre-Output Mixup regularizer, we must have $H\ge C-1$. It should be reminded that $H$ is in fact the number of output neurons for $f^*$. This result is in agreement with the analysis of the Manifold Mixup technique presented by \cite{verma2018manifold}, where the latter only considers non-recurrent frameworks.
\item
Second, the hidden state subsets $\boldsymbol{\mathcal{H}}_j\triangleq\left\{\boldsymbol{h}_{t,i}\vert\boldsymbol{y}_{t,i}=\boldsymbol{e}_j\right\}$ for $j\in\left[C\right]$ must lie in $C$ separate and orthogonal linear subspaces $\mathcal{S}_j\subseteq\mathbb{R}^H$. In other words, we have $\boldsymbol{\mathcal{H}}_j\subseteq\mathcal{S}_j$, and $\mathcal{S}_i\perp\mathcal{S}_j$ for $i\neq j$. We also have
\begin{equation}
\sum_{j=1}^{C}\mathrm{dim}\left(\mathcal{S}_j\right)=H-C+1.
\end{equation}
\end{itemize}
In this regard, the neural network function $f^*$ acts as a state-changing machine that switches among orthogonal subspaces in the hidden-state space $\mathbb{R}^H$ according to the input feature vector $\boldsymbol{x}$. More specifically, assume the machine is at state $\boldsymbol{h}_{t-1}$ and has produced the label $\boldsymbol{y}_{t-1}\in\left\{\boldsymbol{e}_1,\ldots,\boldsymbol{e}_C\right\}$. Therefore, the state vector $\boldsymbol{h}_{t-1}$ must lie in the low-dimensional subspace $\mathcal{S}_{t-1}\subseteq\mathbb{R}^{H}$ which is uniquely determined by $\boldsymbol{y}_{t-1}$ through the procedure described earlier (we have $\boldsymbol{y}_{t-1}=g\left(\boldsymbol{h}_{t-1}\right)$, where $g$ is a linear function). At time step $t$, the input vector $\boldsymbol{x}_t$ is applied to the machine and its state would be changed to $\boldsymbol{h}_t$ which lies on the low-dimensional subspace $\mathcal{S}_{t}$, again uniquely characterized by the label at time $t$, shown as $\boldsymbol{y}_t$.

Interestingly, we have $\boldsymbol{h}_t\in\mathcal{S}_t=\psi\left(\boldsymbol{y}_t\right)$ for all $t\in\mathcal{T}$ where $\psi$ is a fixed mapping from the set of all possible one-hot labels $\left\{\boldsymbol{e}_1,\ldots,\boldsymbol{e}_C\right\}$ to a set of $C$ low-dimensional subspaces in $\mathbb{R}^H$, i.e. $\left\{\mathcal{S}_1,\ldots,\mathcal{S}_C\right\}$. That means the subspace encompassing the hidden state $\boldsymbol{h}_t$ does not depend on the history of the sequence and is uniquely identified by the current label $\boldsymbol{y}_t$. However, if we have $\mathrm{dim}\left(\mathcal{S}_t\right)>0$, then $\boldsymbol{h}_t$ can still take infinitely many values and thus would be able to store information about the past. Therefore, in order to provide the architecture with capability of storing information about the history of a sequence rather than just the current labels, the following inequality must hold:
\begin{equation}
\sum_{j=1}^{C}\mathrm{dim}\left(\mathcal{S}_j\right)=H-C+1\ge\sum_{j=1}^{C}1=C,
\end{equation}
which means one should have $H\ge 2C-1$.

\end{document}